%% file: 0-main.tex
\pdfoutput=1
\documentclass[11pt]{article}
\usepackage{times}
\usepackage{booktabs}
\usepackage{caption}
\usepackage{subcaption}
\usepackage{listings}
\usepackage{latexsym}
\usepackage[normalem]{ulem}
\usepackage{subcaption}
\usepackage{graphicx} 
\usepackage{subfig}   
\usepackage{float}    
\usepackage{dblfloatfix}  
\usepackage{amsmath}

\usepackage{makecell}
\usepackage[utf8]{inputenc}
\usepackage{tcolorbox}
\usepackage{hyperref}
\usepackage[capitalize]{cleveref}
\usepackage{caption}
\definecolor{ballblue}{rgb}{0.13, 0.67, 0.8}
\usepackage[normalem]{ulem}
\hypersetup{
    colorlinks=true,
    linkcolor=ballblue,
    filecolor=ballblue,      
    urlcolor=ballblue,
    citecolor=ballblue,
}
 \newcommand{\robotemoji}{\includegraphics[height=1em]{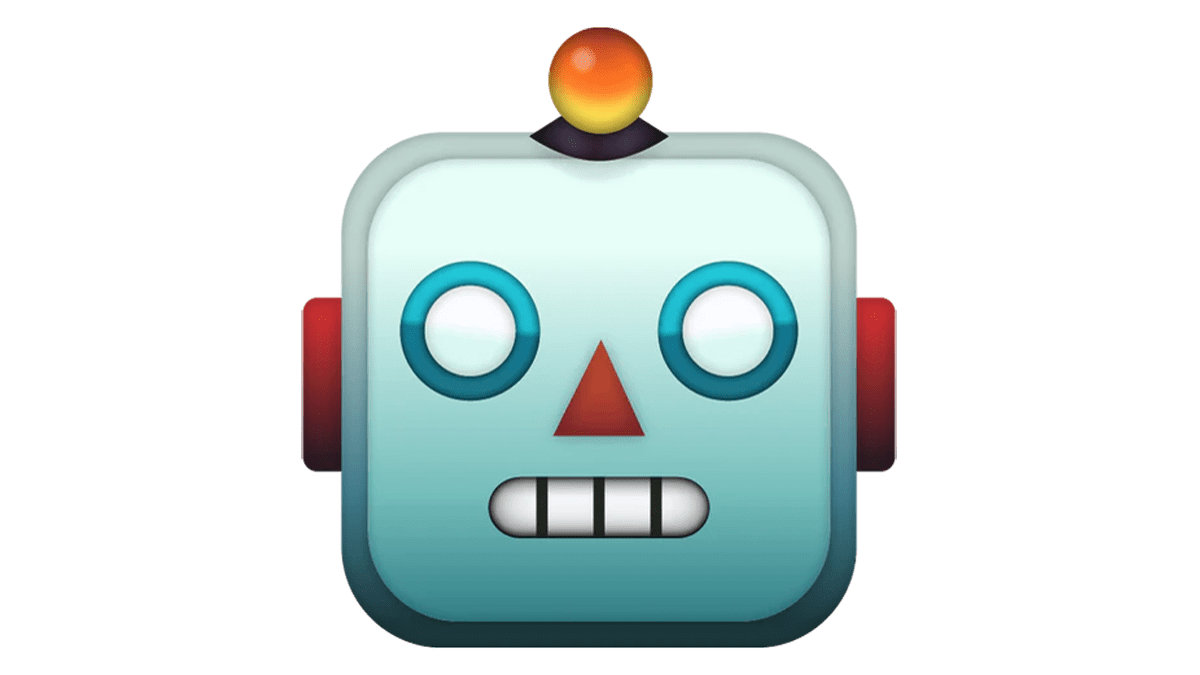}}
\newcommand{\music}[0]{MuSiQue}
\newcommand{\wiki}[0]{2WMHQA}
\newcommand{\wikiori}[0]{2WikiMultiHopQA}

\usepackage{tikz}
\usetikzlibrary{decorations.pathreplacing, positioning}

\usepackage{blkarray}%
\usepackage{xcolor}

\usepackage{ACL_NEW}

\usepackage{lipsum} 

\usepackage[T1]{fontenc}

\usepackage[utf8]{inputenc}

\usepackage{microtype}

\usepackage{inconsolata}

\usepackage{graphicx}
\usepackage{wrapfig}     
\usepackage[dvipsnames]{xcolor}

\usepackage{graphicx}
\usepackage{geometry}
\usepackage{float}
\usepackage{lipsum}  
\usepackage[absolute,overlay]{textpos}

\makeatletter
\@ifundefined{sf@counterlist}{\let\sf@counterlist\@empty}{}
\makeatother

\begin{document}
\title{More Documents, Same Length:\\
Isolating the Challenge of Multiple Documents in RAG}





\author{
 \textbf{Shahar Levy}\footnotemark[1] \qquad
 \textbf{Nir Mazor}\footnotemark[1] \qquad
 \textbf{Lihi Shalmon}\footnotemark[1] \\ 
 \textbf{Michael Hassid} \qquad
 \textbf{Gabriel Stanovsky} \\ 
 School of Computer Science and Engineering\\
 The Hebrew University of Jerusalem,  Jerusalem, Israel \\
 \{shahar.levy2, nir.mazor, lihi.shalmon, michael.hassid, gabriel.stanovsky\}@mail.huji.ac.il
}

\maketitle
\renewcommand{\thefootnote}{\fnsymbol{footnote}}
\footnotetext[1]{Equal contribution.}  
\renewcommand{\thefootnote}{\arabic{footnote}}
\begin{abstract}
Retrieval-Augmented Generation (RAG) enhances the accuracy of Large Language Model (LLM) responses by leveraging relevant external documents during generation. Although previous studies noted that retrieving many documents can degrade performance, they did not isolate how the quantity of documents affects performance while controlling for context length.
We evaluate various language models on custom datasets derived from a multi-hop QA task. We keep the context length and position of relevant information constant while varying the number of documents, and find that increasing the document count in RAG settings poses significant challenges for most LLMs, reducing performance by up to 20\%. However, Qwen2.5 maintained consistent results across increasing document counts, indicating better multi-document handling capability. Finally, our results indicate that processing multiple documents is a separate challenge from handling long contexts. We also make the datasets and code available\footnote{\href{https://github.com/shaharl6000/MoreDocsSameLen}{https://github.com/shaharl6000/MoreDocsSameLen}} to facilitate further research in multi-document retrieval.

\end{abstract}

\input{sections/1-Introduction}

\input{sections/2-Data}
\input{sections/3-Methods}
\input{sections/4-Results}

\input{sections/5-Discussion}




\bibliography{bib}

\appendix
\input{sections/6-DataSet}

\input{sections/7-additional_analysis}

\input{sections/8-prompt}

\input{tables/table}
\end{document}

%% file: sections/1-Introduction.tex
\section{Introduction}

\begin{figure}[tb!]
    \centering
    \includegraphics[width=\linewidth]{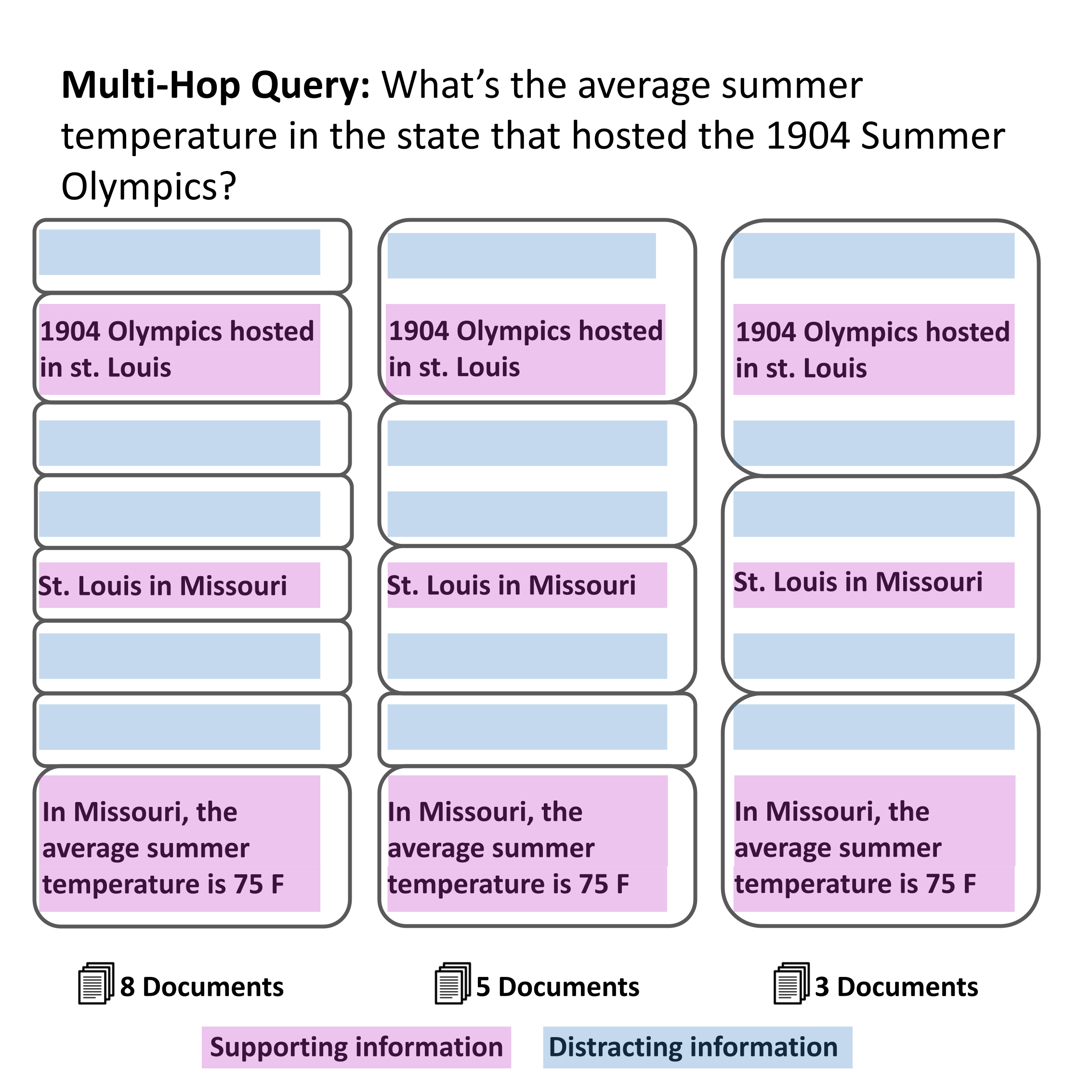}
    \caption{ We create various sets containing the same questions but differing in the number of distractor documents. Each set includes a multi-hop question, all of the supporting documents that contain the information to answer the question (pink), and varying distractor documents (blue). We begin with full-document version (left) and then reduce the number of documents while maintaining a fixed context size. When fewer documents are used, the remaining documents are  extended so that concatenating them yields the same total length.}
    \label{fig:sets}
\end{figure}
The RAG approach enriches prompts with relevant documents, retrieved according to an input query~\cite{karpukhin-etal-2020-dense}. For example, given a question about a certain historical period, RAG techniques can retrieve documents related to the time from a large historical corpus.

Recent work has noted a drop in RAG performance when retrieving many documents. For example, in multi-hop QA, LLMs struggle when the number of retrieved documents grows, even when presented with all the needed information~\cite{press2022measuring, Liu2023LostIT,levy2024tasktokensimpactinput, wang2024leavedocumentbehindbenchmarking}. Such deficiencies were observed without controlling for the \emph{number of tokens} in which the information is conveyed, i.e., when the number of documents grew, so did the number of overall tokens, thus conflating between the challenge of long context and multi document.

In this work, we address the following question: \emph{Assuming a fixed input length, how is LLM performance affected by the number of retrieved documents?}
This disentangles the challenge of long context from the challenge in processing collections of related documents -- which often contain redundancies, 
conflicting information, and implicit inter-document relations~\citep{hirsch-etal-2023-revisiting, lior2024seamstochasticbenchmarkmultidocument}. From a practical perspective, answering this question can help understand a breadth versus depth tradeoff --- i.e., whether to strive to retrieve shorter context out of many documents or whether to aim to retrieve longer context out of fewer documents. An ideal experimental setup would have the exact information conveyed \emph{in the same number of tokens} across varying number of documents, from a long and self-contained single document to a large, multi-document corpus.
We find that the custom sets we constructed from \music{}~\citep{trivedi2022musiquemultihopquestionssinglehop}  and \wikiori{}\citep{ho2020constructing}, a multi-hop QA datasets, serve as a convenient approximation, allowing us to explore the relationship between long-context  and multi-document comprehension in a controlled environment with real-world texts.

Each instance in both datasets consists of a question and a set of documents, where each document is an excerpt from a Wikipedia article retrieved according to the input question. Each instance is constructed such that the question can be answered based on only a subset of the input documents, while the other documents serve as realistic distractors in retrieval settings, as they revolve around the question's topic but do not contain information required to answer the question.

As illustrated in \cref{fig:sets}, we vary the \emph{number of documents} in the input by gradually removing  distractor documents. When removing a distractor document, we respectively extend each of the remaining documents with distracting content from their corresponding Wikipedia article. Importantly, the process preserves the position of the relevant information within the context. 

If the context length is the sole challenge, we should expect the performance to remain similar regardless of the number of input documents. Conversely, if processing multiple related documents presents an additional challenge, we would expect an inverse correlation between performance and the number of input documents.

\begin{figure*}[t!]
    \centering
    \begin{subfigure}{\linewidth}
        \centering
        \includegraphics[width=\linewidth]{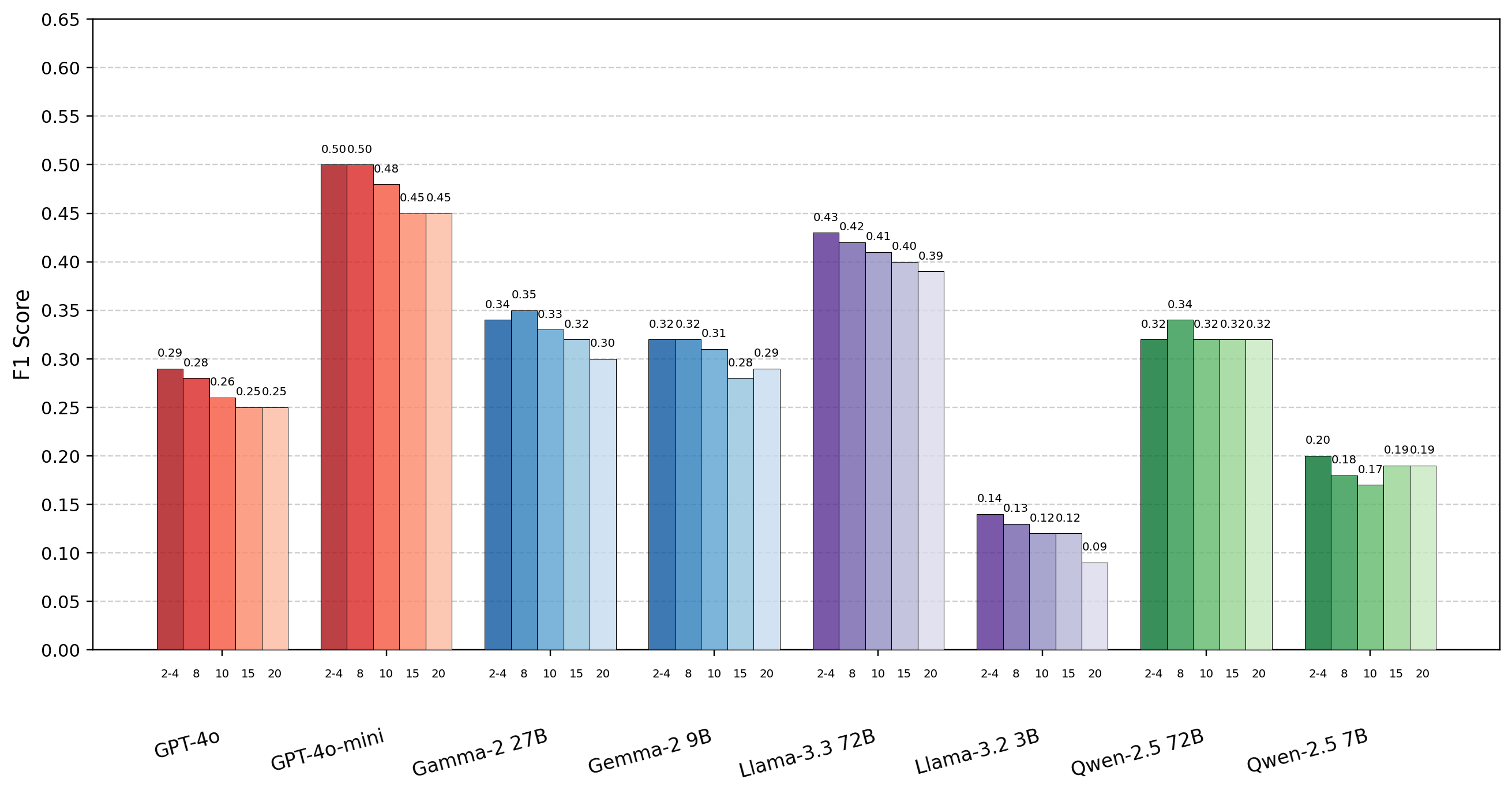}
        \caption{MusiQue}
        \label{fig:musique_subfig}
    \end{subfigure}\\
    \begin{subfigure}{\linewidth}
        \centering
        \includegraphics[width=\linewidth]{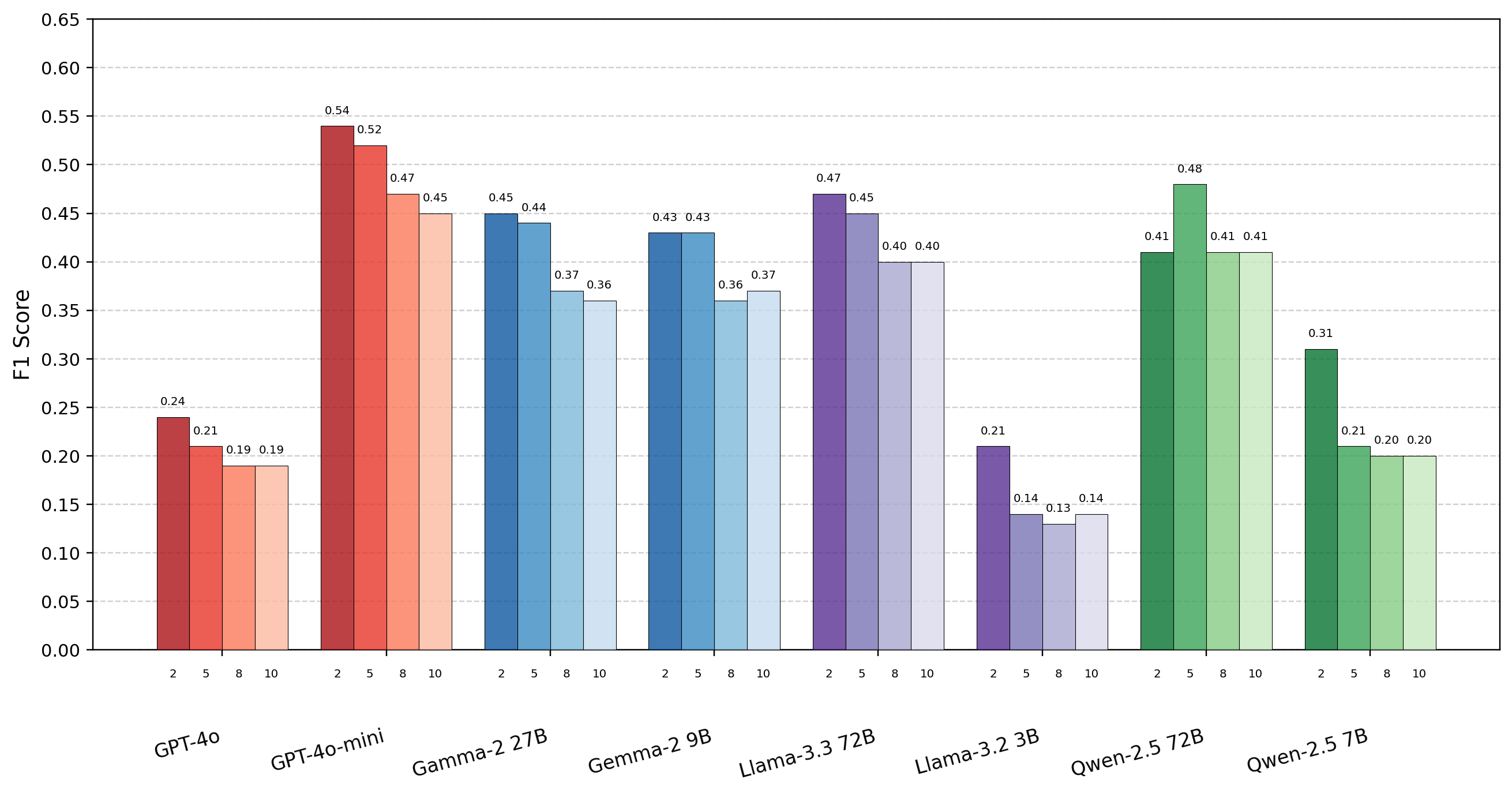}
        \caption{2WikiMultihopQA}
        \label{fig:2wiki_subfig}
    \end{subfigure}
    \caption{
        \textbf{Increasing the number of retrieved documents can hurt performance.}
        In retrieval setups with fixed context windows, adding more documents could reduce performance by up to 10 percent. Two models (Llama-3.3 and Gemma-2) showed worse performance, while Qwen-2.5 remained unaffected. The smaller versions of the LLMs (7–9B) show a similar trend as their larger counterparts but the effect is weaker. The hues of the bars represent the amount of retrieved documents.
    }
    \label{fig:performance_n_docs}
\end{figure*}

Our evaluation of several state-of-the-art models (Llama-3.3, Qwen2.5, Gemma2, and GPT-4o), presented in \cref{fig:performance_n_docs}, indicates that in most cases, reducing the number of documents while keeping the amount of tokens improves performance by up to 10\% in \music{},and up to 20\% in \wiki{}. An exception is Qwen2.5, which may indicate that it better handles multi-document collections.

Our work has several major implications and avenues for future work.
First, from a practical perspective, RAG systems should take the number of retrieved documents into consideration, as the introduction of additional documents into the prompt may hurt performance. Second, future work should explore novel approaches for multi-document processing, which according to our findings presents a separate challenge from mere long context. Such work can make use of our paradigm and data for training and evaluation.

%% file: sections/2-Data.tex




\section{Multi-Document Evaluation with Controlled Token Count}

Our goal is to understand how the number of retrieved documents affects LLM performance when controlling the input length. To this end, we evaluate several models on multi-document multi-hop question answering, which requires models to find relevant information within a given context to answer a specific question. In particular, we make controlled adjustments to the number of documents in the input, while preserving the position of the key information needed to answer the questions, and keeping the context length consistent.


Our datasets are based on  \music{}~\citep{trivedi2022musiquemultihopquestionssinglehop} and \wikiori{}~\citep{ho2020constructing}, which we nickname as \wiki. Both datasets are multi-hop QA datasets that consist of questions associated with paragraphs (20/10 paragraphs for \music{}/\wiki) sampled from individual documents, retrieved from Wikipedia according to the question. Of these paragraphs, 2–4 contain the supporting information necessary to answer the question, while the remaining paragraphs serve as realistic distractors in a RAG setup, as they are retrieved from related topics but do not contain relevant information to answer the question. \cref{fig:sets} shows an example query, and a list of retrieved documents, where three are relevant to the question (marked in pink), and the rest are distractors (marked in blue). We further elaborate on the dataset in section \ref{datasets} in the appendix.
 



Leveraging \music{}’s and \wiki's structure, we constructed several data partitions to investigate the impact of the number of retrieved documents in a controlled manner. The process involves the following three steps:

\paragraph{(1) Selecting the total number of documents:} We reduce the number of documents from the original document count down to only the supporting documents. For \music{} from 20 to 15, then 10, 8, and finally down to the 2--4 documents consisting of the relevant information to answer the question. Similarly for \wiki{}, from 10 to 8, 4, and finally down to the 2 positive documents.

\paragraph{(2) Choosing the supporting and non-supporting documents:} We always keep the documents that support the answer to ensure that the question remains answerable, and randomly select the remaining ones from the non-supporting set. Non-supporting documents remain consistent across different document counts, i.e., each set includes all documents from the smaller sets. \cref{fig:sets} shows such document selection in the two right columns, note that relevant documents (blue) are always kept.

\paragraph{(3) Expanding the selected documents:} Since the original documents consist of Wikipedia paragraphs, we locate their source Wikipedia pages and add text preceding and following the paragraphs to match the original token count. 
Importantly, we uniformly expand supporting and non-supporting documents, under the assumption that both introduce \emph{distracting content}, i.e., content which is not inherently needed to answer the question.  In contrast, \citet{bianchi2025losthaystacksmallerneedles} recently distinguishes between these two types of documents, and considers additional context from supporting documents as beneficial towards answering multi-hop questions. 
Future work may make finer-grained distinctions between document contexts by judging whether their contents help or distract with respect to individual questions.

%% file: sections/4-Results.tex
\renewcommand{\thefootnote}{\arabic{footnote}}


    

\section{Evaluation}

\subsection{Experimental Setup}

We evaluated six instruction-tuned LLMs from four model families: Llama-3.3 70B and Llama 3.2 3B ~\citep{llama3modelcard} \footnote{A small counterpart to Llama-3.3 was not available at the time of evaluation.}, Qwen2.5 7B/72B~\citep{qwen2025qwen25technicalreport}, Gemma2 9B/27B~\citep{team2024gemma}, and GPT-4o/GPT-4o-mini~\citep{hurst2024gpt}. Large models were run on Together.ai\footnote{\href{https://www.together.ai/}{https://www.together.ai}}, and smaller ones on an A6000 GPU. We used a decoding temperature of 0.8, as recommended in prior evaluations~\citep{chen2021evaluating}. Evaluation relied on overlap F1 between gold and predicted outputs, following MuSiQue~\citep{trivedi2022musiquemultihopquestionssinglehop}. Prompts, formats, and evaluation code were implemented using SEAM~\citep{lior2024seamstochasticbenchmarkmultidocument} (see Appendix~\ref{prompt_sec} for details).




\subsection{Results}
Our key findings (\cref{fig:performance_n_docs}) reveal that in a retrieval setup, 
LLMs suffer when presented with more documents, even when the total context length is the same. This may be due to the unique challenges in multi-document processing, which involves processing information that is spread across multiple sources, which can introduce conflicting or overlapping details. 
Almost all models perform better when presented with fewer documents, with scores improving by 5\% to 10\% on average in \music{} and by 10\% to 20\% in \wiki{}. 
We find that the smaller versions of all LLMs exhibit a similar pattern, albeit to a lesser degree.

An exception is Qwen2.5, which may indicate that it better handles multi-document collections. It performed similarly across the different document quantities in \music{} and \wiki{}.

Interestingly, GPT-4o performed significantly worse than GPT-4o-mini. Recent studies show GPT-4o-mini can outperform GPT-4o on certain tasks \cite{chen2024steering, nguyen2025leveraging, alabbasi2025teleoracle}. This may be because GPT-4o's larger parameter count leads to overfitting, while GPT-4o-mini's smaller size forces it to focus on more generalizable patterns.



\subsection{Analysis}
To contextualize our results, we created additional versions of our data, discussed below along with the respective findings.

\begin{table}[t]
    \centering
    \scalebox{0.7}{  
        \begin{tabular}{l cccc}
            \hline
            {\textbf{Model}} & \multicolumn{2}{c}{\textbf{No documents}} \\
            \cmidrule(lr){2-3} 
            & \textbf{\music{}} & \textbf{\wikiori{}}  \\
            \hline
            Qwen-2.5 72B   & 0.01 & 0.03 \\
            Qwen-2.5 7B   & 0.01 & 0.02 \\
            \hline
            Llama-3.3 70B  & 0.05 & 0.08 \\
            Llama-3.2 3B   & 0.01 & 0.01 \\
            \hline
            Gemma-2 27B  & 0.02 & 0.02 \\
            Gemma-2 9B   & 0.05 & 0.02 \\
            \hline
            GPT-4o  & 0.02 & 0.04 \\
            GPT-4o-mini  & 0.05 & 0.01 \\
            \hline
        \end{tabular}
    }
    \caption{F1 scores in a scenario where only the questions are provided (without documents)}
    \label{tab:no_documents}
\end{table}
\paragraph{Contamination does not appear to affect our results.}
To test whether the models relied on memorization, we evaluated them using only the questions, without any retrieved context. All models performed poorly ($\approx$ 0.02 F1), reducing concerns about data contamination. Results in Table~\ref{tab:no_documents}.
\paragraph{Behavior is similar across instances with different amounts of tokens.}
\label{amount_of_tokens}
We evaluate the performance for instances with different context lengths. Although we keep the number of tokens constant across the different multiplicities of documents, each question and its associated documents have a different token count. To further explore whether there is any difference in performance for instances with different lengths, we check the performance as the number of documents increases for different token bins (each bin describes a different range of number of tokens). We observe that for different token bins, the behavior remains the same: as the number of documents increases, the performance degrades. We elaborate in the section \ref{amount_of_tokens} in the appendix.

%% file: sections/5-Discussion.tex
\section{Conclusions}
We assess the challenges of multi-document retrieval tasks when varying the number of documents. Our results indicate that input that includes more documents complicates the task in an environment of retrieval settings, highlighting the need for retrieval systems to balance relevance and diversity to minimize conflicts. Future models could benefit from mechanisms to identify and discard conflicting information while leveraging document variety.

\section{Limitations}
This study does not address prompt variations or the effects of data order within inputs. Future work should explore alternative datasets to ensure more robust evaluations. While our experiments focused on extreme scenarios (highly distracting or random contexts) and document counts between 2–20, future research should investigate more nuanced setups and larger document sets to better reflect real-world conditions. All datasets from this study will be publicly available upon publication for further research in multi-document processing.

%% file: sections/6-DataSet.tex
\section{Datasets}
\label{datasets}
We use two Multi-Hop QA datasets: \music{}~\citep{trivedi2022musiquemultihopquestionssinglehop} and \wikiori{}~\citep{ho2020constructing}, which we nickname as \wiki{}. Both datasets consist of a set of questions associated with documents mined from Wikipedia. The documents are split between those that contain relevant knowledge to solve the question and distractors that contain similar details but not knowledge that is directly relevant to answering the question.

\music{}~\citep{trivedi2022musiquemultihopquestionssinglehop}, a multi-hop QA dataset whose validation set consists of 2,417 answerable questions. Each question is associated with 20 paragraphs sampled from individual documents, retrieved from Wikipedia according to the question. Of these paragraphs, 2–4 contain the supporting information necessary to answer the question, while the remaining paragraphs serve as realistic distractors in a RAG setup, as they are retrieved from related topics but do not contain relevant information to answer the question. The mean token count for questions with their associated documents is 2,400 tokens per instance.

Similarly, \wiki{} is a multi-hop QA dataset which is composed of questions with associated documents, where only a subset are relevant to answering the question. \wiki{}'s validation set consists of 12,576 answerable questions. Differently from \music{}, each question is associated with only 10 paragraphs sampled from individual documents retrieved from Wikipedia. Of these paragraphs, only 2 contain the supporting information necessary to answer the question, while the rest are distractors. In our setup, we choose questions with associated documents that are above 1,500 tokens, which yields a final set of 994 questions. The mean token count for an instance in this dataset is 1,845 tokens.

%% file: sections/7-additional_analysis.tex
\section{Additional Analysis}
\label{additional_analysis}


\begin{table}[t]
    \centering
    \scalebox{0.7}{  
        \begin{tabular}{l cc}
            \hline
            {\textbf{Model}} & \multicolumn{2}{c}{\makecell{\textbf{Supporting} \\ \textbf{documents only}}}  \\
            \cmidrule(lr){2-3} 
            & \textbf{\music{}} & \textbf{\wikiori{}} \\
            \hline
            Qwen-2.5 72B & 0.45 & 0.51  \\
            Qwen-2.5 7B  & 0.23 & 0.29  \\
            \hline
            Llama-3.3 70B & 0.54 & 0.61  \\
            Llama-3.2 3B & 0.15 & 0.20 \\
            \hline
            Gemma-2 27B & 0.52 & 0.57 \\
            Gemma-2 9B & 0.50 & 0.53 \\
            \hline
            GPT-4o & 0.35 & 0.20 \\
            GPT-4o-mini & 0.62 & 0.65 \\
            \hline
        \end{tabular}
    }
    \caption{F1 scores for the large and small versions of each model in the scenario only the supporting documents are provided (without expanding the context). }
    \label{tab:results_supporting_doc}
\end{table}


\begin{table}[t]
    \centering
    \scalebox{0.7}{  
        \begin{tabular}{l ccccc}
            \hline
            {\textbf{Token Bin}} & 2-4 Docs & 8 Docs & 10 Docs & 15 Docs & 20 Docs \\
            \hline
            0 - 2000  & 0.45 & 0.44 & 0.45 & 0.41 & 0.40   \\
            2000 - 2500   & 0.38 & 0.38 & 0.35 & 0.32 & 0.31    \\
            2500 - 3000 & 0.37 & 0.35 & 0.35 & 0.30 & 0.28   \\
            3000+  & 0.30 & 0.29 & 0.26 & 0.28 & 0.28  \\
            \hline
        \end{tabular}
    }
    \caption{F1 scores for Llama-3.1 with 70B parameters performance on \music{}. We clustered the different instances in \music{} according to the number of tokens. We observe that the model's performance degraded as we increased the number of documents, a pattern that occurred across the different bins.}
    \label{tab:bin_count}
\end{table}




\begin{figure*}[t!]
    \centering
    \begin{subfigure}{\linewidth}
        \centering
        \includegraphics[width=\linewidth]{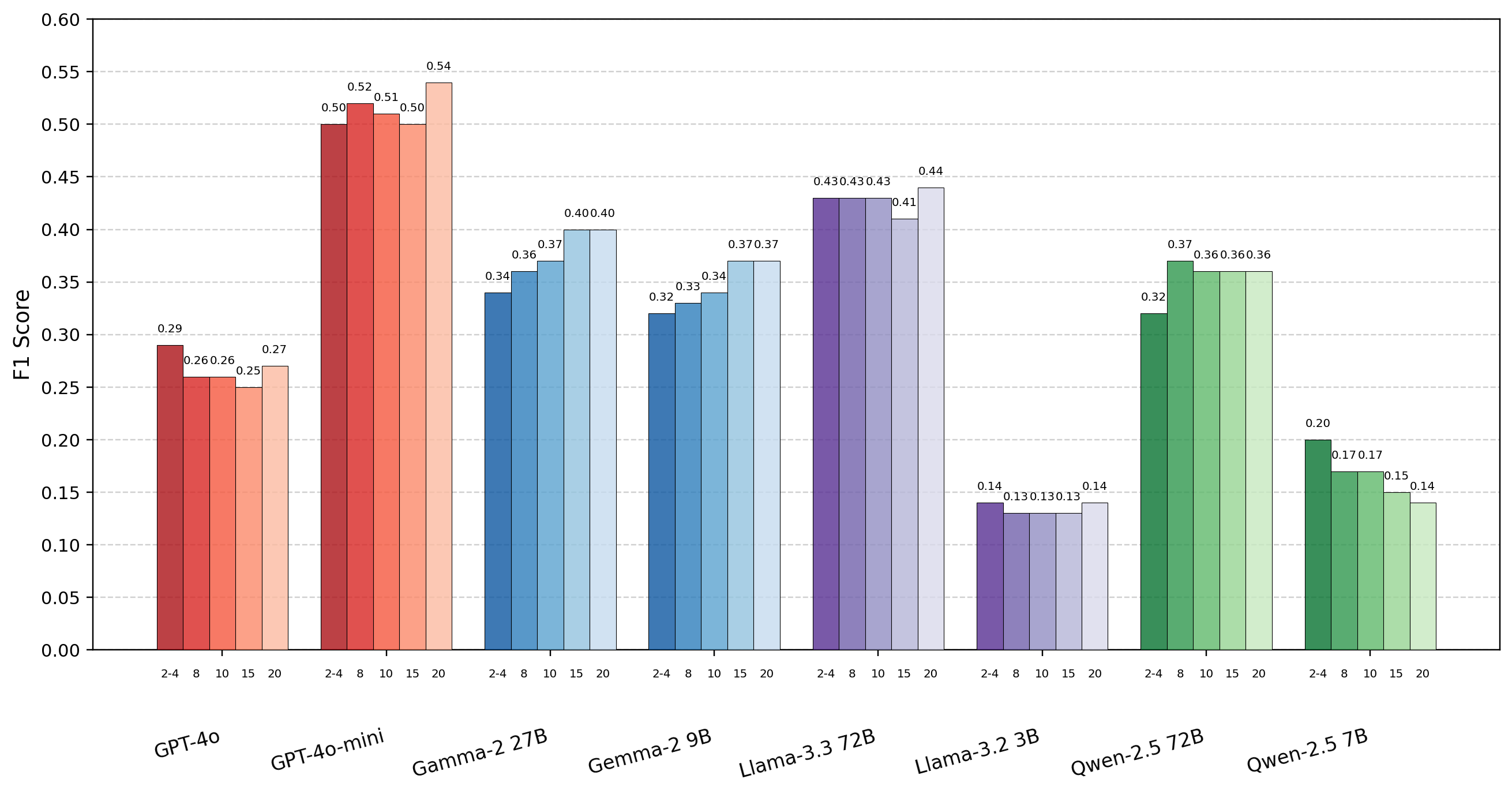}
        \caption{MusiQue}
        \label{fig:musique_subfig}
    \end{subfigure}\\
    \begin{subfigure}{\linewidth}
        \centering
        \includegraphics[width=\linewidth]{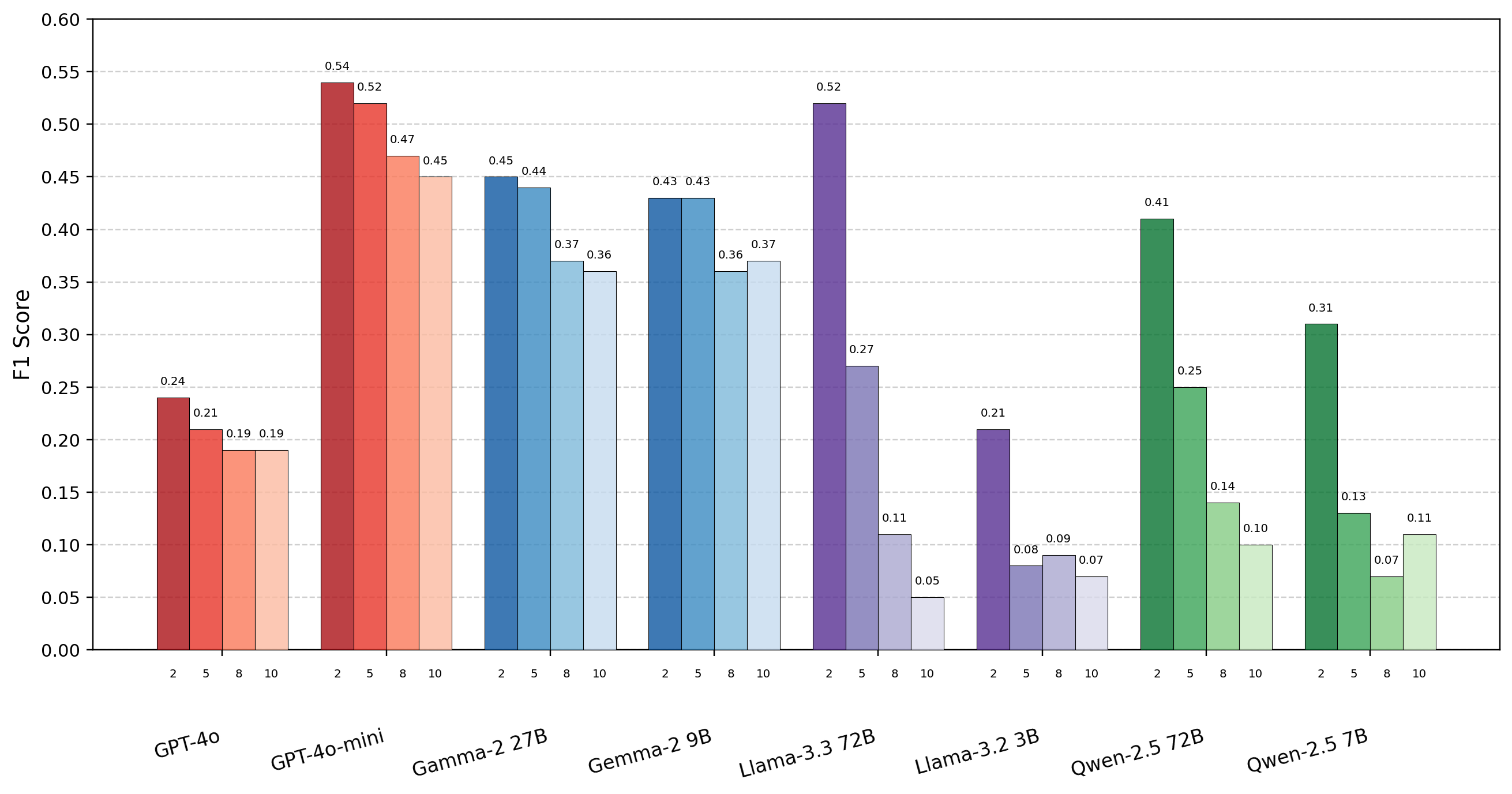}
        \caption{2WikiMultihopQA}
        \label{fig:2wiki_subfig}
    \end{subfigure}
    \caption{
        \textbf{The effects of adding non-related documents.}
    When adding irrelevant documents, LLMs' performance improves across models for \music{} while for \wiki{} it produces significant degradation in performance. 
    }
    \label{fig:random-docs}
\end{figure*}

\subsection{Random distractors yield inconsistent behavior.}
\label{random_distractors}
We evaluate all models against versions of the two datasets where we use randomly selected Wikipedia paragraphs instead of retrieved distractors. As shown in \cref{fig:random-docs}, unlike with the original dataset, we observe more nuanced phenomena. For \music{} with the large LLM versions, the models' performance improves as more documents with random distractors appear within the input. However, for \wiki{}, the performance degrades significantly as more random distractors are added. This suggests that the models' behavior varies significantly when using random distractors compared to retrieved ones. 

We believe the main reason for the difference in performance between the datasets lies in the positive document length: the retrieval content, although it does not contain the answer, still contains relevant information for answering the question. A positive document in MusiQue is around ten sentences, while a positive document in 2WikiMultihopQA contains only two sentences. Since the positive document is shorter, the model may benefit more from additional knowledge from retrieved documents, even if they do not include the actual answer. Therefore, for MusiQue with longer evidence, the model is less reliant on distracting content, while in 2WikiMultihopQA, where the positive document is only two sentences, the knowledge in the retrieved documents might be crucial.

\subsection{Additional context hurts performance.}
\label{baseline}
We test the performance when models are given only the supporting documents, thus providing a much shorter context and eliminating any distracting content. The performance of the LLMs on this set was significantly higher compared to the experimental sets that contained external information. Full results are shown in Table \ref{tab:results_supporting_doc} in the appendix.

\subsection{Observed pattern applies to additional variants}
\label{old_variants}
\begin{figure}[t!]
    \centering
    \begin{subfigure}{\linewidth}
        \centering
        \includegraphics[width=\linewidth]{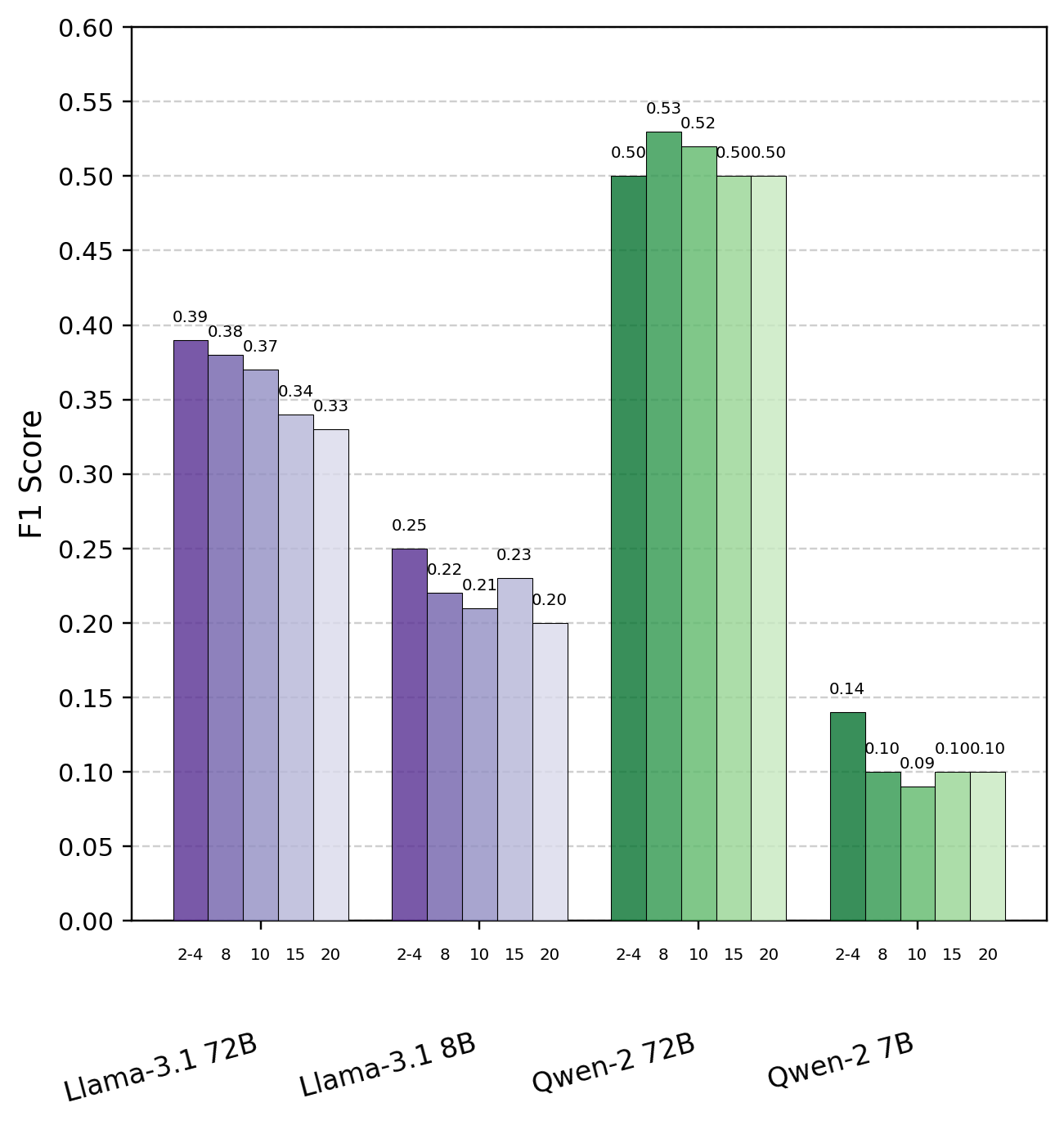}
        \caption{MusiQue}
        \label{fig:musique_subfig}
    \end{subfigure}\\
    \begin{subfigure}{\linewidth}
        \centering
        \includegraphics[width=\linewidth]{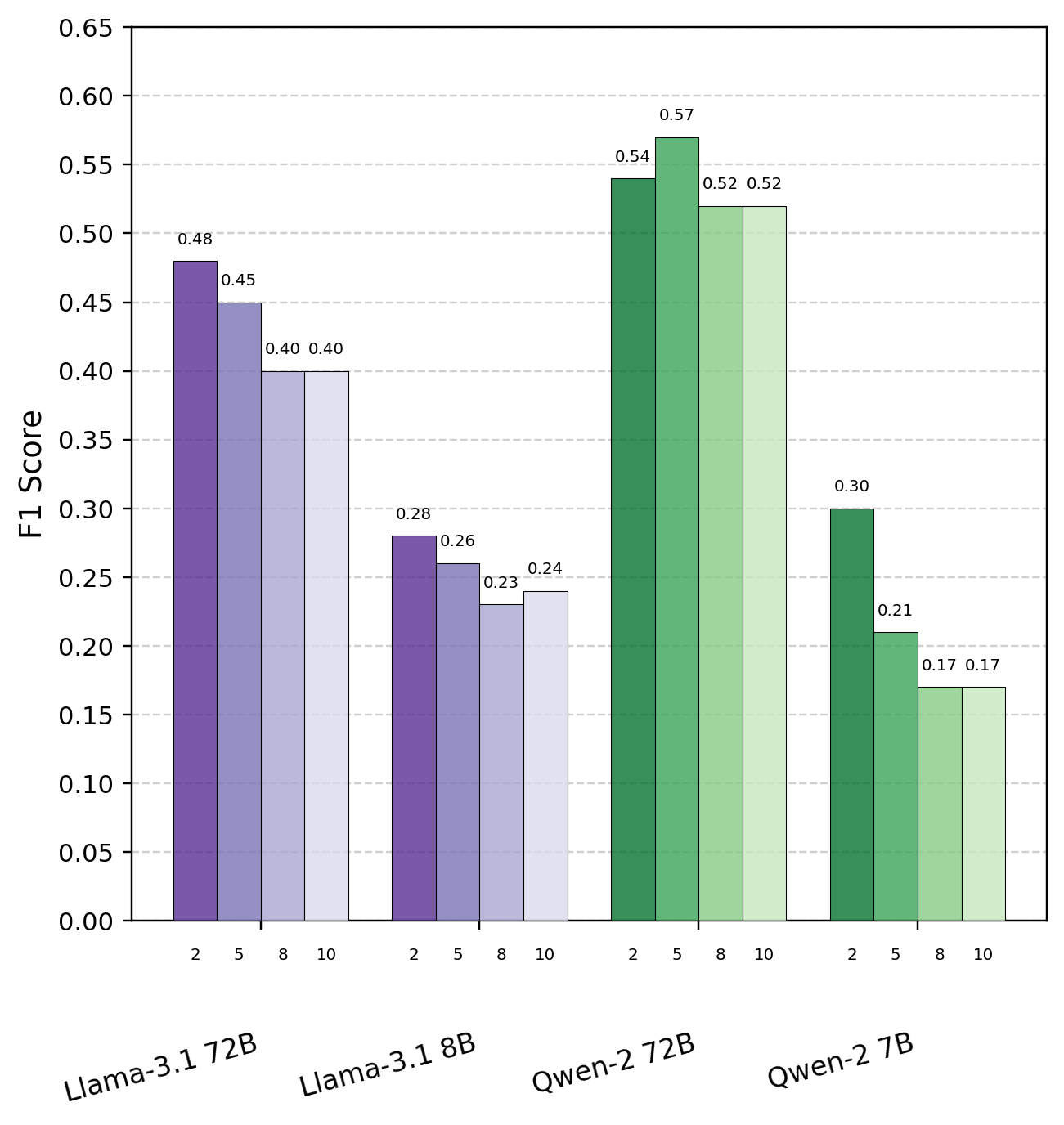}
        \caption{2WikiMultihopQA}
        \label{fig:2wiki_subfig}
    \end{subfigure}
    \caption{
    \textbf{Performance of previous model variants with increasing retrieved documents.}
    We tested earlier model versions (Llama-3.1 and Qwen-2) in retrieval settings with fixed context windows while adding more documents. Our findings were consistent with the latest model versions. Llama-3.1 showed performance reductions of up to 10\%, similar to Llama-3.3, while Qwen-2 remained unaffected, consistent with Qwen-2.5's behavior.
}
    \label{fig:performance_n_docs_old_variants}
\end{figure}

We experimented with two additional model variants: Qwen-2 7B/72B\cite{yang2024qwen2} and Llama-3.1 8B/72B\cite{llama3modelcard}. The observed trends remain consistent across different model versions. Qwen-2 demonstrates robustness as document count increases, suggesting it is better suited for multi-document processing, while Llama-3.1 shows a 10\% performance decrease, as seen in Figure \ref{fig:performance_n_docs_old_variants}.

\begin{figure}[t!]
    \centering
    \begin{subfigure}{\linewidth}
        \centering
        \includegraphics[width=\linewidth]{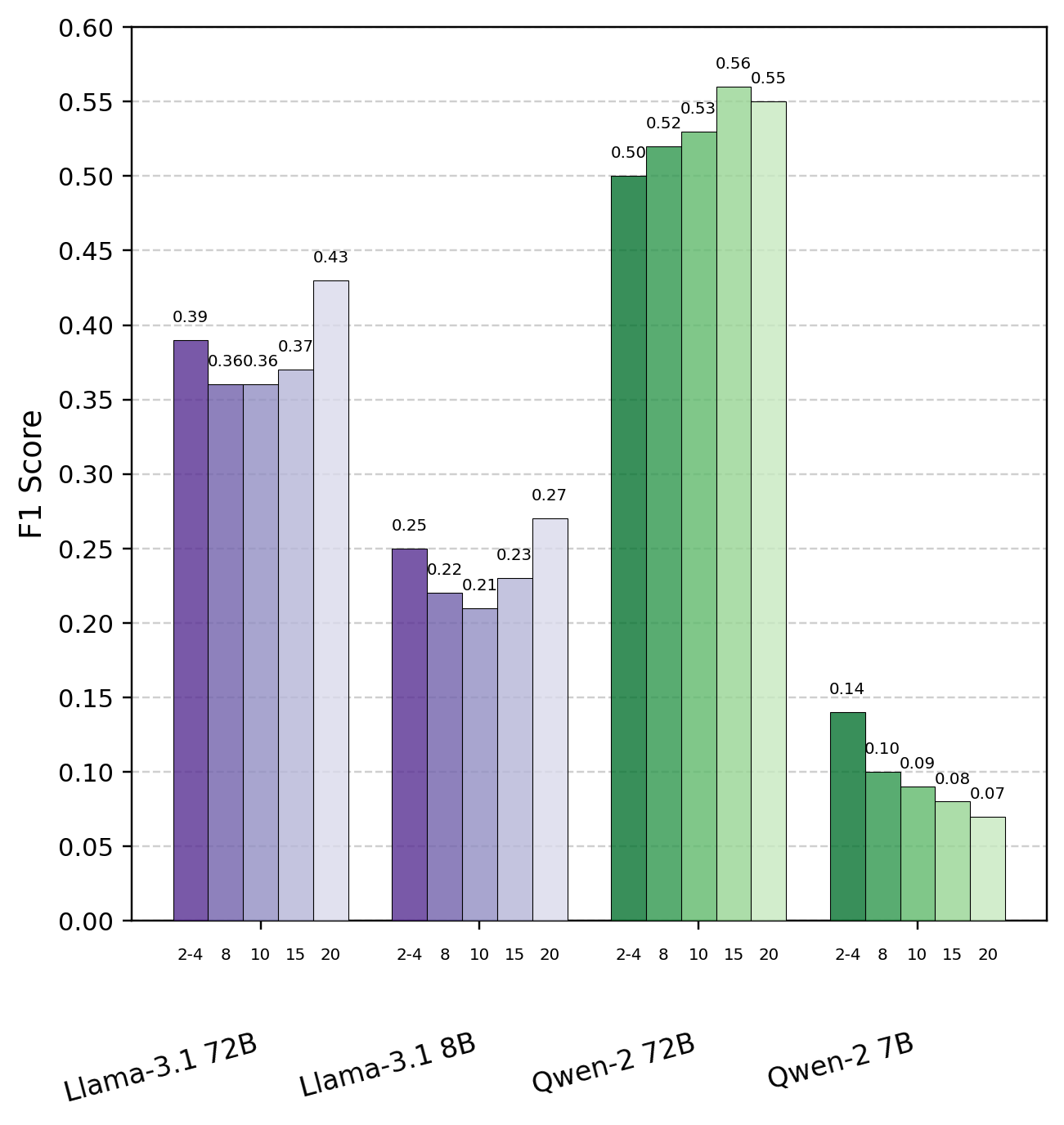}
        \caption{MusiQue}
        \label{fig:musique_subfig}
    \end{subfigure}\\
    \begin{subfigure}{\linewidth}
        \centering
        \includegraphics[width=\linewidth]{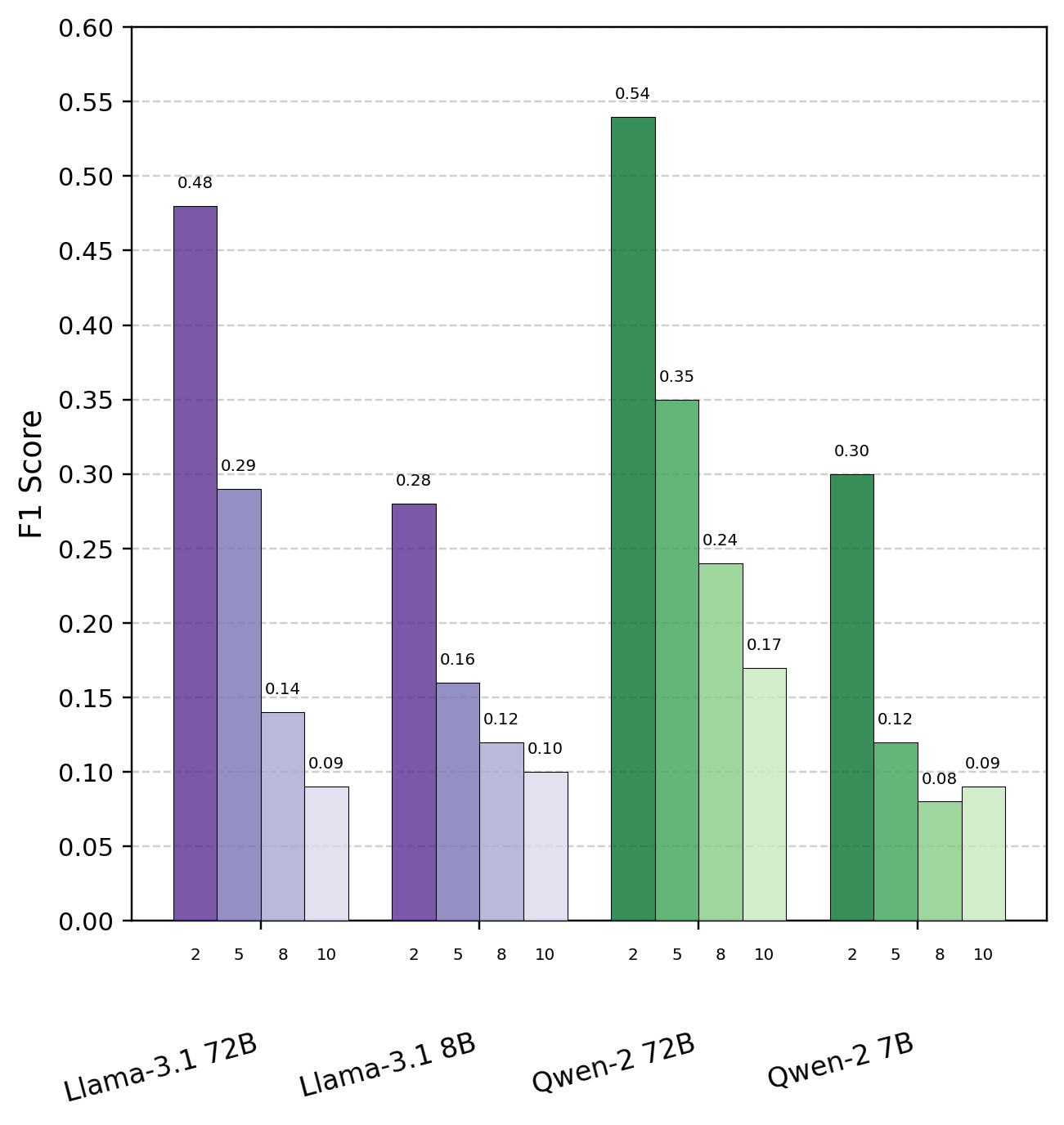}
        \caption{2WikiMultihopQA}
        \label{fig:2wiki_subfig}
    \end{subfigure}
  \caption{
    \textbf{The effects of adding non-related documents for previous variants.}
    Similarly to the latest variants, when adding irrelevant documents, the LLMs' performance improves across models for \music{} while for \wikiori{} it produces significant degradation in performance. 
}
    \label{fig:random-docs-old-variants}
\end{figure}

We also tested these variants with random distractors. Both Qwen-2 and Llama-3.1 exhibit similar patterns to their advanced counterparts: performance improves on \music{} with random documents, while results on \wiki{} degrade significantly as document count increases. Results for random distractors can be seen in Figure \ref{fig:random-docs-old-variants}.

\subsection{Behavior is similar across instances with different amounts of tokens}
\label{amount_of_tokens}
We evaluate the performance of the models for instances with different context lengths. Although we keep the number of tokens constant across the different multiplicities of documents, each question and its associated documents have a different token count. To further explore whether there is any difference in performance for instances with different lengths, we cluster the predictions of Llama-3.1 with 70B parameters on the \music{} dataset according to their number of tokens. Then we check the model performance of each cluster across different numbers of documents. We observe that for each token cluster, the performance still degrades independently of the token count. In addition, we observe that the performance is higher when the number of tokens is lower. Results are presented in Table \ref{tab:bin_count} in the appendix.

%% file: sections/8-prompt.tex
\section{prompt}
\label{prompt_sec}
We use prompt \ref{lst:o1prompt} for all model and document quantities taken from SEAM \citep{lior2024seamstochasticbenchmarkmultidocument}. 

\lstset{
    basicstyle=\ttfamily\small,
    breaklines=true,
    frame=single,
    numberstyle=\tiny,
    captionpos=b,
    keepspaces=true,
    columns=flexible,
    breakatwhitespace=false,
    tabsize=3,
    float=H,         
    aboveskip=10pt,  
    belowskip=10pt,  
    abovecaptionskip=10pt,
    belowcaptionskip=10pt
}

\begin{figure}[t!]
\begin{samepage}
\begin{lstlisting}[caption={Inference prompt \robotemoji}, label={lst:o1prompt}]

In this task, you are presented with a question and 20 documents that contain information related to the question. Your goal is to deduce your answer solely from the provided documents. You must not use any external data sources or prior knowledge.

- Carefully read and analyze each document.
- Identify relevant information to accurately answer the question.
- Formulate a short, concise, and precise answer.
- Exclude irrelevant details from your answer.

Output format:
Return your answer in the following JSON dictionary structure:

- If the provided documents contain the answer:
{
  "is_answerable": true,
  "answer_content": "Your concise answer derived directly from the documents."
}

- If the provided documents do NOT contain sufficient information to answer the question:
{
  "is_answerable": false
}

Important:
- Ensure that your answer strictly adheres to the information in the provided documents.
- Do not include speculation, external facts, or personal interpretations.

\end{lstlisting}
\end{samepage}
\end{figure}